\documentclass[10pt,journal,compsoc]{IEEEtran}
\newcommand{\ie}{\textit{i.e.} }
\newcommand{\eg}{\textit{e.g.} }
\usepackage{verbatim}
\usepackage{pifont}
\newcommand{\cmark}{\ding{51}}%
\newcommand{\xmark}{\ding{55}}%
\usepackage[hyphens]{url}
\usepackage{multirow}
\usepackage{amsmath, amssymb}
\usepackage{url}

\ifCLASSOPTIONcompsoc
  \usepackage[nocompress]{cite}
\else
  \usepackage{cite}
\fi

\ifCLASSINFOpdf
\usepackage[pdftex]{graphicx}
\graphicspath{{images}}
\DeclareGraphicsExtensions{.pdf,.jpeg,.png}
\else
\fi

\begin{document}

\title{An Overview of Privacy-enhancing Technologies in Biometric Recognition}

\author{Pietro~Melzi,
        Christian~Rathgeb,
        Ruben~Tolosana,
        Ruben~Vera-Rodriguez,
        Christoph~Busch% <-this % stops a space
\IEEEcompsocitemizethanks{\IEEEcompsocthanksitem P. Melzi, R. Tolosana, and R. Vera-Rodriguez were with Biometrics and Data Pattern Analytics (BiDA) Lab, Universidad Autonoma de Madrid, Spain.\protect\\
% E-mail: see http://www.michaelshell.org/contact.html
\IEEEcompsocthanksitem C. Rathgeb, and C. Busch were with da/sec -- Biometrics and Internet Security Research Group, Hochschule Darmstadt, Germany.}

}

\markboth{}
{Melzi \MakeLowercase{\textit{et al.}}: An Overview of Privacy-enhancing Technologies in Biometric Recognition}

\IEEEtitleabstractindextext{%
\begin{abstract}
Privacy-enhancing technologies are technologies that implement fundamental data protection principles. With respect to biometric recognition, different types of privacy-enhancing technologies have been introduced for protecting stored biometric data which are generally classified as sensitive. In this regard, various taxonomies and conceptual categorizations have been proposed and standardization activities have been carried out. However, these efforts have mainly been devoted to certain sub-categories of privacy-enhancing technologies and therefore lack generalization. This work provides an overview of concepts of privacy-enhancing technologies for biometrics in a unified framework. Key aspects and differences between existing concepts are highlighted in detail at each processing step. Fundamental properties and limitations of existing approaches are discussed and related to data protection techniques and principles. Moreover, scenarios and methods for the assessment of privacy-enhancing technologies for biometrics are presented. This paper is meant as a point of entry to the field of biometric data protection and is directed towards experienced researchers as well as non-experts.
\end{abstract}

}

\maketitle

\IEEEdisplaynontitleabstractindextext

\IEEEpeerreviewmaketitle

\IEEEraisesectionheading{\section{Introduction}\label{sec:introduction}}

\IEEEPARstart{P}{rivacy} is a broad concept that specifies the right of individuals to protect their freedom and private life from interference or intrusion. The scope of privacy encompasses several areas, for instance it assumes sociological, economical, and political perspectives \cite{westin_social_2003}, %privacy_clarke
and it has been included in numerous documents that define human rights. %\cite{convention, fundamental_rights, universal, convention108}. 
With the rapid development of technologies related to big data, internet of things, artificial intelligence, and cloud computing, among others, the trend has been to collect more and more personal data throughout the years and applications \cite{roh_survey_2021}. %luong_data_2016
As a consequence, the meaning and scope of privacy has evolved, including the right to obtain control over the collection and use of personal data \cite{belanger_privacy_2011}. In 2016, the \emph{General Data Protection Regulation} (GDPR) has been introduced by the European Union to protect individuals with regard to the processing of their personal data \cite{gdpr}. The GDPR describes some principles that must regulate the use of personal data. They include fairness during processing, adequacy to the original purpose of the IT-system, and protection from unauthorised accesses. % Numerous privacy-enhancing technologies (PETs) have been proposed to address such requirements. 
To address such requirements the GDPR suggests technical approaches like pseudonymisation, %\cite{pfitzmann_anonymity_2001}, 
which refers to processing of personal data in a way that they can no longer be attributed to specific individuals without the use of additional information. Other approaches considered by the GDPR are encryption and data minimisation. In addition, the GDPR incorporates \emph{privacy by design} (PbD), a concept initially developed by Ann Cavoukian, according to which the protection of data must be integrated in a system from its creation to achieve strong privacy protection without diminishing its functionality \cite{cavoukian_privacy_2010}. Furthermore PbD mandates that privacy settings in an IT-system are activated as default. %According to the former, data protection must be integrated in the system from its creation, according to the latter the processing of personal data must be limited to those data that are specific for the purpose of the system. We refer to technical approaches that address such privacy requirements with the term ``privacy-enhancing technologies'' (PETs).
Interestingly, the same author suggested the application of the PbD concept to systems processing biometric data, because of the potential issues of data misuse and security vulnerabilities identified there \cite{cavoukian_biometric_2011}. %cavoukian_advances_2012

Biometric data are measurements of human characteristics with the purpose of recognising and describing individuals. They can be divided in two categories: \emph{i)} biological and \emph{ii)} behavioural \cite{jain_introduction_2004}. Biological data are related to individuals' bodies, such as fingerprints, faces, and irises, while behavioural data involve the individuals' actions (\ie functions of the body), such as keystroke, gait, signature, and speech. Biometric data are widely used in recognition systems as they are unique to each individual and cannot be forgotten, lost, and transferred to other individuals, such that this authentication factor offers a clear advantage over to traditional knowledge- and possession-based authentication systems \cite{bhattacharyya_biometric_2009}. % However, it is important to note that, unlike a password, a compromised biometric data may be revoked and renewed only a limited number of times \cite{ISO-IEC-30136}. For instance, the number of different fingerprints that an individual can provide is limited. 
Biometric recognition systems have been implemented in various application scenarios and with multiple purposes, such as automated border control, access control in high security facilities, e-banking, healthcare, forensic for law enforcement, and smartphone unlocking. %\cite{s22030792, SCHLETT2021103247, TOLOSANA2022108609, 8998358, melzi_analyzing_2021}. % However, they also present critical aspects that require to be addressed, given the widespread use of these systems.
At the time of enrolment, \textit{i.e.} registration, biometric data are usually stored as biometric reference. Biometric data are commonly represented as so-called \emph{templates}, \ie sets of biometric features related to an individual and comparable directly to probe biometric features extracted during authentication. For instance, minutiae extracted from fingerprints \cite{903046}, or binary iris-codes extracted from irises \cite{5276817}. According to \cite{vocabulary_busch}, with the term \emph{biometric data} we refer to both, the original representations of human characteristics as well as the features extracted from them.

\begin{figure*}[tb]
    \centering
    \includegraphics[width=0.8\linewidth]{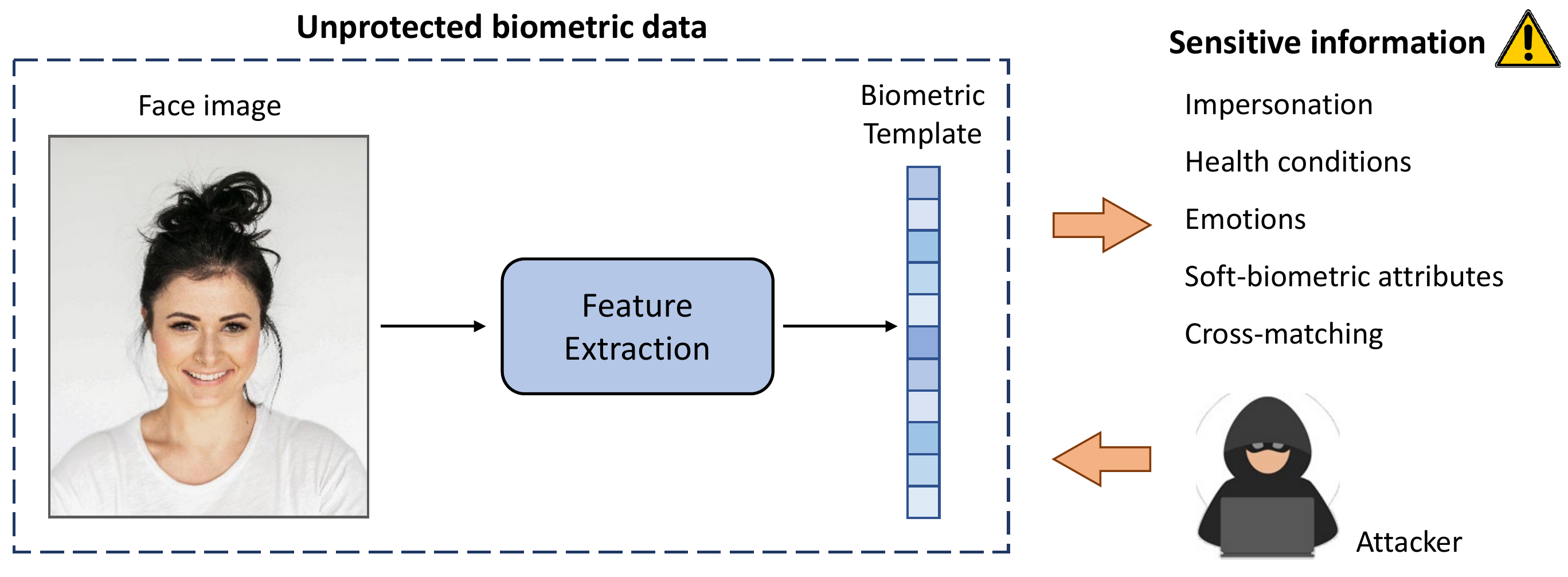}
    \caption{Privacy concerns derived from the storage of unprotected representations of biometric data and unprotected biometric templates. In the figure we consider a face image. Similar concerns apply to other biometric traits.}
    \label{fig:soft_biom}
\end{figure*}
The unprotected storage of biometric data raises privacy concerns about the ultimate use of them. If the original representation (\ie captured samples like face images or fingerprint images) are stored, they could potentially leak out from the server\footnote{https://www.opm.gov/news/releases/2015/09/cyber-statement-923/}. An attacker who compromises a biometric database can obtain the biometric data of the enrolled individuals, and eventually impersonate them to gain access to the corresponding authentication system. In addition, further information can be derived from the biometric data, including health conditions, emotions, soft-biometric attributes, and other personal aspects \cite{delgado-santos_survey_2021, melzi_analyzing_2021}. Also, we note that storing processed biometric data (\eg feature vectors) is not a protection level, as image representations can be easily reconstructed from biometric features in many cases, \textit{e.g.} in \cite{8338413}. %4288153

Soft-biometric attributes consist in information contained in biometric data that increases the chances to recognise individuals. %\cite{dantcheva_what_2016}. 
Soft-biometric attributes, such as age, gender, ethnicity, and many more, can be automatically extracted from biometric data without the user's agreement, and used for purposes that were not originally intended. % Such technologies usually incorporate the concept of data minimisation, as many soft-biometric attributes may be extracted from biometric data and used for purposes that are not the originally intended ones \cite{}. Such attributes are not necessarily unique to individuals, but provide additional information that increases the chances to recognise them \cite{dantcheva_what_2016, gonzalez-sosa_facial_2018}. Classic examples of soft-biometric attributes are: age, gender, and ethnicity. %, height, weight, activity, and health status-related information.
% They can be automatically extracted from biometric data without user's agreement, thereby raising several privacy concerns \cite{mirjalili_soft_2017, terhorst_comprehensive_2021, rot_detecting_2022, delgado-santos_survey_2021}.
% In Figure \ref{fig:face_template}, we indicate some examples of soft-biometric attributes that can be extracted from face images and mobile interaction, revealing sensitive information of individuals. 
Moreover, if multiple systems in which individuals have registered their biometric data are compromised, an attacker can cross-match biometric data across such systems to gain further profiling information about individuals \cite{ratha_enhancing_2001}. In Figure \ref{fig:soft_biom} we summarise the privacy concerns in the case of face images. Similar concerns arise from other types of biometric characteristics. Following the definition of biometric data in \cite{vocabulary_busch}, which reads biometric sample or aggregation of biometric samples at any stage of processing, \eg biometric reference, probe, biometric feature or biometric property, we consider the entire variety as sensitive data, and formulate and review the application of privacy-enhancing technologies (PETs) to them to provide a solution for the privacy concerns discussed above. 

We observe that some of the approaches to privacy enhancement by biometric data protection investigated in the literature cannot be successfully implemented in the scenario of biometric recognition systems. For instance, anonymisation techniques provide strong privacy assurances but prevent the recognition of individuals, eliminating the utility of biometric data. % Similarly, differential privacy is used to aggregate information about individuals while preventing the derivation of information regarding a particular individual. 
Moreover, traditional encryption algorithms % such as AES \cite{} 
cannot be applied to biometric data because %of the \emph{avalanche effect}: 
small changes in the original data, such as the unavoidable variances between multiple biometric data measurements from the same characteristic of the same individual, cause a drastic change in the encrypted data. Hence, encrypted biometric data need to be decrypted before comparing them, with consequent advantages for the attackers \cite{rathgeb_biometric_2017}. Finally, it is important to remark that biometric data require a special effort compared to other data to protect the additional information, for instance related to the health status of the captured subject, that can be easily derived from the data.

In this article, we refer to PETs suitable for biometric recognition systems with the term \emph{biometric privacy-enhancing technologies} (B-PETs), and to the data generated by B-PETs with the term \emph{protected biometric data}. We provide a summary of concepts with practical relevance, so that practitioners can attain a comprehensive overview of B-PET concepts through this work and identify the best approach to achieve privacy enhancement in their application, and according to their needs. Existing surveys in this field do not fulfil this function of guideline and limit their scope to specific (sub-)categories of B-PETs. For instance, cancelable biometrics and biometric cryptosystems are investigated in \cite{nandakumar_biometric_2015, rathgeb_survey_2011}, while the removal or concealing of specific information from biometric data is investigated in \cite{meden_privacyenhancing_2021}. The present article aims to gather under a general framework different B-PETs that are usually investigated in separate works in the literature. In particular in \cite{meden_privacyenhancing_2021}, even if a wide range of B-PETs is described, the scope of the work is limited to technologies designed according to the principle of data minimisation. Moreover, some of the technologies described there aim to prevent the recognition of individuals from biometric data, while maintaining the utility of biometric data in applications that involve soft-biometric attributes, in contrast to our concept of B-PETs. A general taxonomy for PETs has been proposed in \cite{heurix_taxonomy_2015}. Compared to this work, we exclusively consider PETs applied to biometric recognition systems (B-PETs) and focus on the protection of the stored data over transmitted data, as the latter involves network protocols that are not exclusive to biometric data. To sum up, these are the main contributions of our work:
\begin{itemize}
    \item We consider PETs in widespread application of biometric recognition, allowing for a concrete analysis of the different aspects and goals considered for privacy enhancement.
	\item We put focus on a set of well-established categories of B-PETs and investigate them through the definition of a general framework that highlights properties, purposes, and singularities of each category. In this way, the reader can identify the most suitable B-PET for their application, according to the advantages and disadvantages depicted in this overview. 
	\item We qualitatively evaluate existing categories of B-PETs according to their ability to satisfy specific privacy requirements and prevent the extraction of sensitive information from the protected biometric data that they generate. During the evaluation, we consider attackers with different capabilities and knowledge of B-PETs.
\end{itemize}

The remainder of the article is organised as follows. In Section \ref{sec:fundamentals} we introduce fundamentals of biometric recognition systems and B-PETs. Section \ref{sec:general_framework} describes the proposed general framework for B-PETs, providing details of the different categories of B-PETs. % and specifying the privacy requirements that they address. 
Section \ref{sec:evaluation} presents some metrics suitable for the evaluation of B-PETs, and performs an assessment of the considered categories of B-PETs under different attack scenarios. Finally, Section \ref{sec:conclusion} draws conclusions and  points out future research lines in this research field. 

\section{Fundamentals}\label{sec:fundamentals}
In this section, fundamental concepts of biometric recognition systems and B-PETs are introduced to facilitate a full understanding of this work.%\footnote{Readers that are already familiar with the fundamental concepts of biometric recognition systems and B-PETs can skip to the next section.} 

\subsection{Biometric recognition systems}
Biometric recognition systems perform an automated recognition of individuals based on their behavioural and biological characteristics. They are generally composed of four subsystems that allow to capture biometric samples of individuals, process and compare them to determine whether individuals are recognised or not \cite{jain_introduction_2004}. We describe the subsystems in the following. Recall that according to \cite{vocabulary_busch} biometric data is defined as biometric sample or aggregation of biometric samples at any stage of processing:
\begin{itemize}
    \item \emph{Data capture:} it captures biometric samples of individuals through capture devices.
    \item \emph{Signal processing} and \emph{feature extraction:} they processes the captured biometric samples to extract a set of salient or discriminatory features (\ie a feature vector) from them. 
    \item \emph{Comparison:} it performs comparisons between acquired biometric data $x$ and stored biometric data $y$, and generates similarity scores $s = S(x, y)$, according to some similarity functions $S$. Based on similarity scores, the system decides if two biometric feature vectors are from the same subject (match) or from different subjects (non-match).
    \item \emph{Data storage:} it stores the biometric reference (\ie biometric data) of enrolled individuals, and provides them when necessary to perform comparisons.
\end{itemize}

Subsequent presentations leading to multiple representations (\ie biometric data that are either samples or feature vectors) provided by the same individual usually exhibit some variance. Hence, it is essential to choose the proper threshold $t$ to determine if two biometric data $x$ and $y$ belong to the same individual, \ie if $S(x, y) \geq t$. High thresholds may result in false non-matches of mated comparison trials (\ie genuine attempt), while low thresholds may result in false matches of non-mated comparison trials (\ie impostor attempt). According to these error types, common metrics to measure the performance of biometric recognition systems are: \emph{i)} false match rate (FMR), \ie the probability that an impostor is incorrectly accepted as genuine, and \emph{ii)} false non-match rate (FNMR), \ie the probability that a genuine individual is incorrectly rejected as impostor. %, and equal error rate (EER), \ie the value at which FMR and FNMR are equal.

\subsection{Biometric privacy-enhancing technologies}
We have already discussed the privacy concerns related to the collection, processing, and storage of biometric data, and the consequent application of B-PETs to address them. B-PETs generate protected biometric data that should satisfy the privacy requirements specified in Section \ref{sec:privacy_req}. We denote the application of B-PETs as a function $f$ applied to biometric data $x$: $\tilde{x} = f(x, k)$, where $\tilde{x}$ are the resulting protected biometric data, and $k$ represents optional parameters of $f$. We organise the different B-PETs proposed in the literature in the following categories:
\begin{itemize}
    \item \emph{Cancelable biometrics:}  consist of intentional, repeatable distortions of the original biometric data based on transformations which enable a comparison of biometric data in the transformed domain \cite{ratha_enhancing_2001}.
    \item \emph{Biometric cryptosystems:}  are designed to securely bind a digital key to biometric data or generated a digital key from biometric data \cite{cavoukian_biometric_2011}.
    \item \emph{Homomorphic encryption:} allows to generate the encrypted result of operations performed on plaintexts directly computing operations on ciphertexts, \ie without any intermediate decryption \cite{gomez-barrero_multi-biometric_2017}. % \cite{fontaine_survey_2007}.
    \item \emph{Soft-biometric minimisation:} identify soft-biometric attributes in the representations of biometric data, discard them, and generate new representations of biometric data excluding such soft-biometric attributes \cite{bortolato_learning_2020}.
    \item \emph{Soft-biometric protection:}  modify the representation of biometric data to prevent the extraction of soft-biometric attributes \cite{morales_sensitivenets_2021, terhorst_pe-miu_2020, delgado-santos_gaitprivacyon_2021}. In this case, soft-biometric attributes are not discarded, but they are considered inaccessible in the new representations of biometric data. 
\end{itemize}

\begin{figure*}[tb]
    \centering
    \includegraphics[width=0.85\linewidth]{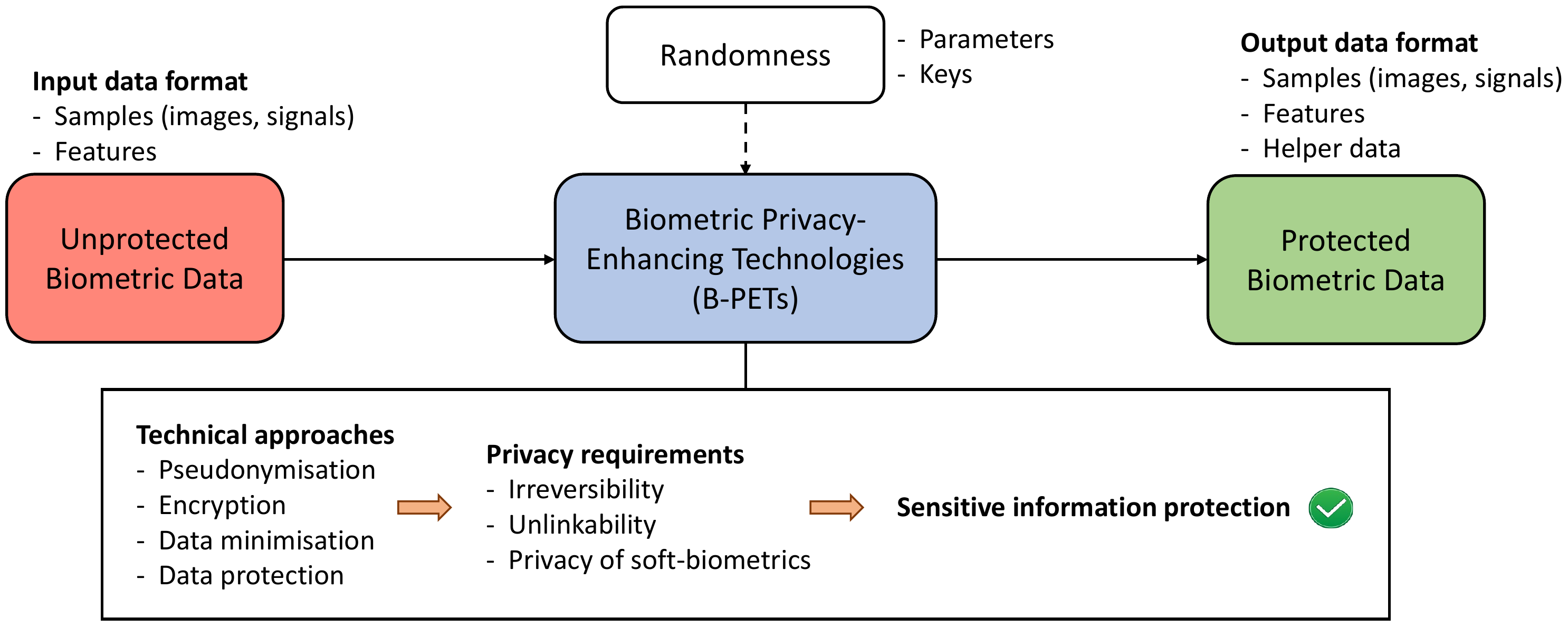}
    \caption{Representation of the general framework and the technical approaches considered to generate protected biometric data. The dashed line indicates an optional component of the framework.}
    \label{fig:framework}
\end{figure*}

\subsection{Privacy requirements}\label{sec:privacy_req}
The protected biometric data obtained from B-PETs are required to satisfy specific privacy requirements to overcome the concerns related to the use of biometric data in recognition systems. Some requirements are long established \cite{nandakumar_biometric_2015}, although not always properly evaluated. More recently, novel threats related to the possible extraction of soft-biometric attributes as well as other information from biometric data emerged \cite{terhorst_beyond_2020}. As a consequence, privacy requirements need to be improved continuously to maintain their effectiveness. In the following, we describe the main privacy requirements for B-PETs \cite{ISO-IEC-24745-TemplateProtection-2022}:

\begin{itemize}
    \item \textbf{Irreversibility:} it should be difficult to reconstruct biometric samples similar to the original captured samples from the stored protected biometric data. Irreversibility can be achieved by applying irreversible transformations or transformations that make use of secret parameters to biometric data. We highlight the importance of B-PETs also considering the possibility of partial irreversibility of protected biometric data. In fact, partial reconstructions of the original data may reveal soft-biometric information of individuals, and may allow attackers to access the system. 
    
    Given a function $g$ that attempts to reconstruct the original biometric data $x$ (\ie a sample) from the protected biometric data $\tilde{x}$, such that $x' = g(\tilde{x})$ is the reconstructed biometric data, irreversibility is achieved if $S(x, x') < t$, for any biometric similarity function $S$ and any given threshold $t$.
    
    \item \textbf{Unlinkability:} it should be difficult to determine if different representations of protected biometric data belong to the same individual or not. Unlinkability can be obtained by introducing some randomness with keys or random parameters in transformations that are %different secret keys or transformations when 
    protecting biometric data. Unlinkability shall prevent cross-matching attacks across multiple systems. When unlinkability is satisfied, compromised biometric data can be revoked and substituted with new protected representations. 
    
    Given the protected biometric data $\tilde{x}_1=f(x, k_1)$, $\tilde{x}_2=f(x, k_2)$, $\tilde{y}_1=f(y, k_1)$, and $\tilde{y}_2=f(y, k_2)$, obtained from B-PET $f$ with different biometric data $x$ and $y$ and different parameters $k_1$ and $k_2$, unlinkability is achieved if $P(S(\tilde{x}_1, \tilde{x}_2) \geq t) = P(S(\tilde{x}_1, \tilde{y}_1) \geq t) = P(S(\tilde{x}_1, \tilde{y}_2) \geq t)$, for any given threshold $t$.
    % \item \textbf{Confidentiality:} biometric data must be protected against unauthorized accesses that may result in privacy risks.
    
    \item \textbf{Privacy of soft-biometrics:} the extraction of soft-biometric attributes from biometric data for purposes different than the originally intended ones must be prevented. We observe that false positive comparisons carried out in biometric recognition systems should not disclose any information about the soft-biometric attributes of protected biometric data, which is usually not the case \cite{osorio-roig_attack_2022}. 
    
    Given a soft-biometric attribute $A$, the set of its possible values $\{a_1, a_2, ..., a_n\}$, and a soft-biometric classifier $h$ trained to determine the soft-biometric attribute $A$ from biometric data $x$, the following condition must be valid to ensure the privacy of soft-biometric attribute $A$: $P(h(\tilde{x}) = a_i) = P(h(\tilde{x}) = a_j), \forall a_i, a_j \in A$.
\end{itemize}

The compliance with these privacy requirements results in a sound protection of biometric data and other information that can be obtained from them. In case of unlinkability, the relationship with the required technical approaches is straightforward: they must incorporate some sort of randomness to provide unlinkability. Differently, more technical approaches can provide irreversibility, with encryption that relies on the strength of the key, and data minimisation that ensures unrecoverable elimination of specific information.

While cancelable biometrics and biometric cryptosystems are usually claimed to provide irreversibility and unlinkability, the fulfilment of these requirements may be overestimated during evaluation. Furthermore, the possibility of extracting soft-biometric attributes from biometric data is usually investigated by training classifiers with unprotected biometric data and evaluating them with protected biometric data. However, if a classifier is not able to learn the pattern of an attribute, this does not imply that the pattern does not exist \cite{terhorst_beyond_2020}. Also, some patterns may differ between the unprotected and protected representations of biometric data. In Section \ref{sec:evaluation} we discuss accurate ways to carry out the evaluation of privacy requirements for the different categories of B-PETs. 

\begin{table*}[]
\centering
\caption{Overview of the different categories of B-PETs, in terms of characteristics of their implementation.}
\label{tab:comparsion1}
\scalebox{1.00}{\renewcommand{\arraystretch}{1.4}%
\begin{tabular}{lcccc}
\textbf{B-PETs}             & \textbf{Input data format} & \textbf{Randomness} & \textbf{Output data format} & \textbf{Comparison} \\ \hline\hline
Cancelable biometrics       &  Samples, features                     &    Parameters                 &    Samples, features                    &    Standard                 \\ \hline
Biometric cryptosystems     &   Features                    &   Keys                  &    Helper data                    &     Key retrieval                \\ \hline
Homomorphic encryption      &   Features                    &   Keys                  &   Encrypted features                     &     In encrypted domain                \\ \hline
Soft-biometric minimisation &    Samples, features                   &     No                &     Samples, features                   &  Standard                   \\ \hline
Soft-biometric protection   &    Samples, features                   &   No                  &     Samples, features                   &    Standard                 \\ \hline
\end{tabular}
}
\end{table*}

\section{General framework}\label{sec:general_framework}
The protection of biometric data is an essential aspect in biometric recognition systems, given the multiple privacy concerns that may arise from the unprotected storage and potential misuse of biometric data. Over the years, many B-PETs have been proposed to enhance the privacy of biometric data and they have been categorised according to certain properties. In this section, we introduce a general framework to outline the different components that constitute classes of B-PETs, showing the common aspects and (dis)similarities between different categories of B-PETs. Figure \ref{fig:framework} provides an illustration of said general framework. %which are usually not considered in the literature.
With the introduction of this framework, we show that the categories of B-PETs are not mutually exclusive, as it may appear from the literature, and that different B-PETs can be combined together to improve the final protection of biometric data.

\subsection{Architecture of the framework }\label{subsec:framework_architecture}
We describe the different components of the general framework and how they can be implemented in the different categories of B-PETs, highlighting the distinctive traits of them. In this sense, Table \ref{tab:comparsion1} provides a comparison of the main characteristics related to the implementation of the considered categories of B-PETs.

\subsubsection{Input data format}
The common procedure in biometric recognition systems consists in capturing biometric data through capture devices and subsequently extracting biometric features from the captured samples, \ie numbers or labels used to compare the representations \cite{vocabulary_busch}. Feature extraction allows to reduce the size of biometric data and map them into a discriminative space, where different representations of individuals can be well separated. %\cite{rida_comprehensive_2020}. 
Hence, B-PETs can be applied to biometric data at different levels:
\begin{itemize} 
    \item \emph{Sample level:} B-PETs are applied directly to the collected images or signals, modifying the appearance of biometric data from a human point of view. Biometric features can be subsequently extracted, and it is assumed that the protection introduced at sample level is transferred to them. 
    \item \emph{Feature level:} B-PETs are applied to the extracted features and relate to the form in which machines observe and process biometric data, trying to prevent the execution of automatic and unintended operations on biometric data \cite{morales_sensitivenets_2021}.
\end{itemize}
% A further distinction between B-PETs applied to features is proposed by \cite{meden_privacyenhancing_2021}, that distinguishes between B-PETs applied during enrolment, \ie at \emph{representation level}, and B-PETs applied during the matching or classification stages of biometric recognition systems, \ie at \emph{inference level}.
% According to \cite{hahn_towards}, in case of face images two main approaches ...

All categories of B-PETs provide solutions working at feature level. This is not the case of sample level, where the bigger size of biometric data may pose a limit for some B-PETs. Nevertheless, several cancelable biometrics technologies have been designed to transform data at sample level \cite{ratha_enhancing_2001}, %zuo_cancelable_2008
and a list of B-PETs directly applied to face images is presented in \cite{hahn_towards_2022}. Lastly, soft-biometric minimisation and protection technologies can be applied at sample level \cite{delgado-santos_gaitprivacyon_2021,meden_privacyenhancing_2021}.

\subsubsection{Randomness}
Randomness is an optional although widely employed component that can be provided as further input to B-PETs in the categories of cancelable biometrics, biometric cryptosystems, and homomorphic encryption. The incorporation of randomness in B-PETs allows the generation of multiple templates from the same biometric data, enhancing the desired properties of unlinkability and renewability required for protected biometric templates. In cancelable biometrics the random element consists of user-specific or application-specific parameters of the transform that must be secretly stored, as they are required during authentication and, if compromised, they may allow attackers to launch \emph{linkage attacks}, as well as facilitating the reconstruction of the original biometric data \cite{rathgeb_survey_2011}. For instance, BioHashing can rely on user-specific randomness generated from a seed stored in USB token or smart card microprocessor \cite{jin_biohashing_2004}. In biometric cryptosystems, explicit random values and additional sources of randomness are combined to biometric data in multiple ways, for instance in fuzzy embedders \cite{buhan_survey_2010}. %buhan_embedding_2008
To provide unlinkability in homomorphic encryption, random numbers can be combined together with biometric data, prior to encrypting them with the same public key \cite{gomez-barrero_multi-biometric_2017}. The encryption key itself is also random.

Achieving unlinkability and renewability without the use of randomness is basically infeasible. B-PETs that apply deep neural networks (DNNs) to transform biometric data in a privacy-protected version, according to some criteria, have been proposed, for instance in \cite{morales_sensitivenets_2021}. Typically, randomness is not included in the architecture of DNNs. An exception is represented by the randomised DNN proposed in \cite{mai_secureface_2021} for face template protection, where two kinds of randomness have been considered: \emph{i)} random activation of DNN neurons and \emph{ii)} random permutation and sign flip of extracted templates. That work provides an analysis showing that the proposed protected template satisfies the criteria of unlinkability.  

\subsubsection{Output data format}
B-PETs are applied to unprotected biometric data to generate protected biometric data that satisfy specific privacy properties and consequently can be stored in biometric recognition systems without arising certain privacy concerns. The format of protected biometric data varies according to the B-PETs considered to generate them and the format of input data. Typically, protected biometric data come in the format of obscured biometric features, and they are obtained when B-PETs are applied to unprotected biometric features \cite{nandakumar_biometric_2015} or directly to original biometric data \cite{pinto_secure_2021}. However, it is also possible that the application of B-PETs to images or signals generates protected biometric data that maintain the same format of the input, as in the case of image morphing \cite{korshunov_using_2013}. %or signal reconstruction through autoencoders \cite{malekzadeh_protecting_2018, boutet_dysan_2021, delgado-santos_gaitprivacyon_2021}.

Differently from the other categories of B-PETs, the output of biometric cryptosystems is generally more complex to describe and assumes the name of \emph{helper data}. It can be obtained according to numerous algorithms from biometric features, eventually combined with a secret key \cite{buhan_survey_2010}. %juels_fuzzy_1999, chang_biometrics-based_2004
Helper data alone should not reveal information about the original biometric data and key. They may as well be unprotected, for instance alignment information could be provided in plain format. % \cite{sood_methods_2014}. % However, when biometric data sufficiently similar to the original ones are provided, helper data allow to recover the secret key or the original biometric features \cite{rathgeb_survey_2011}.

\subsubsection{Data comparison}
Protected biometric data are employed for the recognition of individuals in biometric recognition systems. Individuals provide to the system their probe biometric data, that (in most cases) will be protected with the same B-PET used during enrolment. Recognition can be carried out in two modes: 
\begin{itemize}
    \item \emph{Verification:} individuals also have to claim their identity. The recognition process consists in a single comparison between the probe biometric data and the previously enrolled biometric data that serves as biometric reference for the claimed identity.
    \item \emph{Identification:} the comparison between probe and enrolled biometric data is usually performed for each individual enrolled in the system (exhaustive search). Identification can be seen as a sequence of verifications.
\end{itemize}
For simplicity, in this study we consider the single comparison performed during verification.

The modality of comparison in biometric recognition systems depends on the format assumed by the protected biometric data. If B-PETs do not modify the original format of biometric data, the comparison can be performed in the \emph{transformed domain} with the same comparator of the original unprotected system, as in the case of most cancelable biometrics \cite{patel_cancelable_2015}. Eventually, individuals have to present their secret parameters along with probe biometric data, if they are stored outside the biometric recognition system \cite{jin_biohashing_2004}. It is also possible that some B-PETs randomly change the order of biometric features to protect the information contained therein. Hence, such B-PETs require some operations to reorder biometric features before comparison \cite{terhorst_pe-miu_2020}.

Finally, the comparison can be made in the encrypted domain when biometric data are protected with specific categories of B-PETs that make use of secret keys. In case of biometric cryptosystems, at the time of authentication no secrets need to be presented, but only biometric data. Biometric cryptosystems are usually employed in verification scenarios, and the result of a comparison consists in the disclosure of the key to the individual, or in a failure message. In case of homomorphic encryption, the comparison produces a comparison score that, once decrypted, is identical to what would be obtained if the computation was carried out in the unencrypted domain, at the cost of increased computation and communication overhead.

\subsection{Technical approaches for privacy enhancement}
Privacy enhancement consists in the adoption of measures and precautions during the processing of personal data, to increase their protection according to predefined principles without losing the system functionalities. Privacy enhancement involves numerous privacy-related aspects, as one can observe in privacy regulations, \eg GDPR, where multiple principles related to the processing of personal data have been defined \cite{gdpr}. Privacy-related measures usually encompass reduction of or denial of access to personal data. % \cite{gan_privacy_2019}. 
B-PETs are designed incorporating different technical approaches to provide privacy enhancement by executing suitable operations on biometric data. To be more effective, B-PETs usually focus on specific technical approaches, and can be considered more or less suitable according to the context of application. However, the distinction between existing B-PETs is not sharp, with numerous B-PETs that enhance privacy of biometric data according to multiple privacy-related aspects at the same time, as reported in Table \ref{tab:comparsion2}.

\begin{table}[]
\centering
\caption{Summary of the technical approaches implemented by the different categories of B-PETs. PN = pseudonymisation, EN = encryption, DM = data minimisation, DP = data protection.}
\label{tab:comparsion2}
\scalebox{1.00}{\renewcommand{\arraystretch}{1.4}%
\begin{tabular}{lcccc}
\textbf{B-PETs}             & \textbf{PN} & \textbf{EN} & \textbf{DM} & \textbf{DP} \\ \hline \hline
Cancelable biometrics       &   \cmark          &        \cmark     &           &     \cmark        \\ \hline
Biometric cryptosystems     &   \cmark          &        \cmark     &             &     \cmark      \\ \hline
Homomorphic encryption      &    \cmark         &        \cmark     &             &     \cmark        \\ \hline
Soft-biometric minimisation &             &             &    \cmark         &             \\ \hline
Soft-biometric protection   &             &             &             &    \cmark         \\ \hline
\end{tabular}
}
\end{table}

In the following, we describe some technical approaches implemented by B-PETs to provide privacy enhancement during the processing of biometric data:
\begin{itemize}
    \item \emph{Pseudonymisation:}  consists of the generation and use of pseudonymous identifiers (PI) to identify individuals instead of their real identifiers, such as names and biometric data. A general architecture to obtain PIs from biometric data is described in the ISO/IEC 24745 \cite{ISO-IEC-24745-TemplateProtection-2022}. Differently from \emph{anonymisation} which is completely irreversible and not suitable for the applications considered in this study, pseudonymisation allows to uniquely identify particular individuals while hiding their actual identity \cite{heurix_taxonomy_2015}. Additional information is required to attribute pseudonyms to specific individuals. %This information is usually contained in private data and provides the relationship between pseudonyms and real identifiers. The access to additional information is critical to effectively achieve privacy enhancement.
    \item \emph{Encryption:} the encoding of human readable information into a coded format that can only be interpreted by authorised parties, preventing unauthorized access to data. Encryption and decryption of data are performed with public algorithms that rely on secret keys. As we have previously noted, traditional encryption is not applicable to biometric data. However, specific encryption algorithms and suitable techniques to address the biometric variance can be successfully implemented, for instance homomorphic encryption \cite{gomez-barrero_multi-biometric_2017}. %drozdowski_application_2019    
    \item \emph{Data minimisation:} follows a principle also considered in the GDPR, according to which the processing of personal data should be limited to what is necessary to achieve the purposes of the system \cite{gdpr}. To address data minimisation, B-PETs identify the sensitive information included in biometric data and not necessary for the system of interest, and discard it. The extraction of features from biometric data can be considered as preliminary approach to data minimisation: the reconstruction of original biometric data may be difficult in some cases, but the sensitive information is still present in biometric features.
    \item \emph{Data protection:} this term generally assumes wide and inaccurate meanings. In this study we consider as data protection any type of data transformation intended to prevent the extraction of information from biometric data. Differently from data minimisation where certain information is separated and discarded from biometric data, in data protection, all information remains included in biometric data, but it is considered inaccessible.
\end{itemize}

Cancelable biometrics and biometric cryptosystems generate protected biometric data that do not reveal significant information about the original data or the identity of their owner \cite{rane_standardization_2014}. The recognition process carried out with these protected biometric data is considered fully pseudonymous, as the original biometric data are never exposed during comparisons \cite{rathgeb_survey_2011}. This is also the case of homomorphic encryption, with comparisons performed in the encrypted domain. Hence, we observe that there is a close relationship between pseudonymisation and encryption. When pseudonymisation is achieved through encryption, the privacy assurances completely rely on the security of the encryption process. Otherwise, in cancelable biometrics pseudonymisation can be obtained with parameterised irreversible transformations that provide different privacy assurances, but biometric data are still processed in a protected domain. Finally, biometric cryptosystems implement cryptographic algorithms with some error tolerance in order to generate protected biometric data. % while also achieving data minimisation \cite{cavoukian2007biometric}.

We distinguish some categories of B-PETs that are specifically designed to prevent the extraction of soft-biometric attributes from biometric data, differently from cancelable biometrics that protect the overall information contained in biometric data given the computational difficulty to recover the original biometric data from the transformed one \cite{terhorst_pe-miu_2020}. These B-PETs are not originally intended to pseudonymise or encrypt biometric data, with such characteristics that possibly result as a consequence of data transformations. On the contrary, the implementations of these B-PETs follow the approaches of data minimisation and data protection. According to the former, sensitive information is identified in biometric data, and a novel representation of biometric data that does not embed such sensitive information is finally learned \cite{terhorst_suppressing_2019}. According to the latter, the representation of biometric data is modified so that the extraction of sensitive information is made impossible in the novel representation \cite{terhorst_pe-miu_2020}.

\subsection{Additional requirements}
Together with privacy requirements, B-PETs are required to satisfy additional requirements, given that privacy enhancement schemes should not affect the functionality of the system according to the privacy-by-design concept. Hence, even if not directly addressing privacy-related aspects, additional requirements are equally important to obtain practical privacy enhancement in biometric recognition systems. We describe such requirements in the following:

\begin{itemize}
    \item \emph{Minimisation of accuracy degradation:} the performance of biometric recognition systems provided with unprotected biometric data should be maintained when processing protected biometric data.
    %\item \emph{Renewability:} when protected biometric data are compromised, it should be possible to revoke and substitute them with new protected biometric data obtained from the same individual.
    \item \emph{Computational requirements:} the processing of protected biometric data should not cause a significant increase of computational costs in the system, compared to the original processing of unprotected biometric data.
    \item \emph{Storage requirements:} protected biometric data should not significantly increase storage requirements. 
    % \item \emph{No selectivity:} privacy enhancement is not limited to specific soft-biometric attributes, usually specified before the application of B-PETs.
\end{itemize}

\section{Evaluation of the framework}\label{sec:evaluation}
A standardised evaluation of B-PETs is difficult to achieve, as numerous B-PETs have been proposed in the literature and attacks are specifically developed to target the different B-PETs. For this reason, the ISO/IEC 30136 introduces general concepts and metrics for the evaluation of privacy requirements at an abstract level, while concrete evaluations are specific to the architectures considered \cite{rane_standardization_2014, ISO-IEC-30136}. In this section, we discuss important aspects to consider for the evaluation of privacy requirements. We observe that a comprehensive evaluation should also account for specific attacks that may compromise B-PETs without being included by common metrics during the evaluation of privacy requirements. 

\begin{table*}[]
\centering
\caption{Comparison of privacy requirements for the different categories of B-PETs when \textbf{standard model} and \textbf{full-disclosure model} are considered for attacks. We do not consider the advanced model, as the executable submodules are specific for each B-PET. Parenthesis indicate that the validity/non-validity of requirements depends on the goodness of the reconstructed biometric data. SB = soft-biometrics.}
\label{tab:requirements}
\scalebox{1.00}{\renewcommand{\arraystretch}{1.4}%
\begin{tabular}{@{\extracolsep{6pt}}lcccccc@{}}
\multicolumn{1}{c}{\multirow{2}{*}{\textbf{B-PETs}}} & \multicolumn{3}{c}{\textbf{Standard model}}                                                                                                                                                                              & \multicolumn{3}{c}{\textbf{Full-disclosure model}}                                                                                                                                                                          \\ \cline{2-4} \cline{5-7} 
\multicolumn{1}{c}{}                                 & \textbf{Irreversibility} & \textbf{Unlinkability} & \textbf{Privacy of SB} & \textbf{Irreversibility} & \textbf{Unlinkability} & \textbf{Privacy of SB} \\ \hline \hline
Cancelable biometrics       &    \cmark                                                                     &  \cmark                                                                     &         (\cmark)                                                               &         (\xmark)                                                                  &     (\xmark)                                                                       &     (\xmark)                                                                                    \\ \hline
Biometric cryptosystems     &     \cmark                                                                    &    \cmark                                                                   &              \cmark                                                          &      \xmark                                                                    &             \xmark                                                           &   \xmark                                                                               \\ \hline
Homomorphic encryption      &      \cmark                                                                   &       \cmark                                                                & \cmark                                                                       &      \xmark                                                                    &     \xmark                                                                   &      \xmark                                                                                                                             \\ \hline
Soft-biometric minimisation &      (\xmark)                                                                   &  \xmark                                                                     & \cmark                                                                       &      (\xmark)                                                                    &       \xmark                                                                 &         \cmark            \\ \hline
Soft-biometric protection   &                                 (\xmark)                                                                   &  \xmark                                                                     & \cmark                                                                       &      (\xmark)                                                                    &       \xmark                                                                 &         \cmark                                      \\ \hline
\end{tabular}
}
\end{table*}

\subsection{Setup}
The evaluation of B-PETs can be carried out in two ways:
\begin{itemize}
    \item \emph{Theoretical}: it consists in a formal demonstration of the attack potential, or the advantage of an attacker over random guessing. It considers \emph{information-theoretic} metrics related to entropy \cite{nandakumar_biometric_2015}, and relies on statistical assumptions that may, however, result in an overestimation of privacy properties. 
    \item \emph{Empirical}: it consists in assessing the feasibility of implemented attacks in terms of computational complexity \cite{nagar_biometric_2010}. Indeed, even if protected biometric data expose no information at all, an attacker can use brute force to guess the original data.
\end{itemize}
Theoretical and empirical evaluations complement each other: the former usually shows whether an algorithm has potential vulnerabilities, while the latter shows if attackers can exploit them \cite{Zhou2012PrivacyAS}. 

\subsubsection{Threat models}
The evaluation of B-PETs requires the specification of threat models, to represent the expertise and a priori information at disposal of attackers. The following threat models have been described in the ISO/IEC 30136 \cite{ISO-IEC-30136}:
\begin{itemize}
    \item \emph{Naive model:} the attacker has neither information about the algorithms implemented by B-PETs, nor owns a large biometric database. They have only access to the attacked protected biometric data. We do not consider meaningful the evaluation of B-PETs according to this threat model.
    % \item \textbf{Collision model}: The attacker has a large amount of biometric data and uses them to exploit inaccuracies in biometric systems. It can find in its own database biometric data that generate protected biometric templates sufficiently similar to those of an individual enrolled in the system.
    \item \emph{General model:} the attacker knows the algorithms implemented by B-PETs, the statistical properties of biometric features, and has access to the protected biometric data. Privacy protection relies on the presence of secret parameters, from which further threat models are built upon each other:
    \begin{itemize}
        \item \emph{Standard model:} the attacker cannot execute the submodules that make use of secrets.
        \item \emph{Advanced model:} the attacker can execute part of the submodules that make use of secrets.
        \item \emph{Full-disclosure model:} all the secrets are disclosed to the attacker, that can execute the entire system.
    \end{itemize}
\end{itemize}
The different categories of B-PETs are usually evaluated according to the standard and full-disclosure models, where the only difference between the two models consists in the knowledge of the secrets by the attacker. In Table \ref{tab:requirements} we provide an overview of the evaluation of B-PETs that we describe in Section \ref{sec:eval_pr_req}. While specific implementations of B-PETs may fail to satisfy privacy requirements, the table refers to the purposes that categories of B-PETs are expected to achieve, according to the two threat models.

\subsection{Evaluation of privacy requirements} \label{sec:eval_pr_req}
In this section we discuss multiple aspects to consider to effectively evaluate if privacy requirements are satisfied by B-PETs. Together with threat models, it is important to regard the target of attackers, as it can be many things (Figure \ref{fig:soft_biom}). When reconstructions of biometric data are considered \emph{sufficiently good}, it depends on the target of the attack. For instance, the target can be to achieve similarity scores above the recognition threshold, or to derive the soft-biometric attributes of the original biometric data.

\subsubsection{Irreversibility}

Numerous information-theoretic metrics have been proposed to measure the irreversibility of protected biometric data, for instance \emph{conditional entropy} $H(x|\tilde{x})$ quantifies the uncertainty in estimating the original data $x$ from the protected biometric data $\tilde{x}$ \cite{nandakumar_biometric_2015}. As metrics of entropy are difficult to compute theoretically, irreversibility is usually measured empirically according to the computational complexity of the best-known inversion attack. However, the attacker may come up with a better attack not known by the system designer \cite{nandakumar_biometric_2015}. \emph{Privacy leakage} is proposed to evaluate irreversibility in \cite{rane_standardization_2014} with the number of bits leaked about the original data when (part of) the protected biometric data is compromised. In this sense, privacy leakage can also assess the partial reconstruction of biometric data, that may lead to successful attacks even if the original data is not completely reversed. 

To evaluate irreversibility we consider an analogy with cryptography, where the security level of an algorithm is expressed in \emph{bits}, with \emph{n}-bit security meaning that the attacker must perform $2^n$ operations to break the system. However, the same evaluation does not directly apply to B-PETs. While cryptographic systems require exact inputs to provide the desired outputs, in biometric systems similar enough approximations of biometric data may suffice. Also, compared to cryptography, an educated guess based on the statistics of biometric features can facilitate the reconstruction of the original biometric data. This is due to \emph{broad homogeneity}, according to which the biometric data of individuals of the same sex or ethnicity present similar characteristics \cite{howard_effect_2019}. In cryptography a minimum security level of 100-bit is required to consider attacks impractical \cite{aumasson_too_2019}. This means that brute force can always be applied to guess short biometric feature vectors. Finally, we observe that the application of B-PETs at feature level makes the recovering of original data more difficult, as both the reconstruction of biometric features and original data must be done.

Different aspects relate to the evaluation of irreversibility for the different categories of B-PETs. Cancelable biometrics require the analysis of the computational effort to reverse transformations and approximate the original biometric data. Biometric cryptosystems require keys with sufficient size and entropy, so that the number of guesses necessary to retrieve the biometric data or the key itself is high \cite{rathgeb_survey_2011}. In the case of B-PETs that discard or obscure biometric features to prevent the extraction of soft-biometric attributes, sufficiently good approximation of the original biometric data may be obtained even if soft-biometric attributes are protected. %In the case of B-PETs that obscure biometric features to protect soft-biometric attributes, transformations can be reversed in presence of other protected biometric data of the same individual \cite{terhorst_pe-miu_2020}, which is a condition fulfilled even by naive attackers. 
In conclusion, we highlight the difference between B-PETs that apply encryption to protect biometric data, like homomorphic encryption, and B-PETs that apply data minimisation. For the former, irreversibility completely relies on the secrecy of the key: if known, the original data can be immediately obtained. For the latter, irreversibility does not depend on secrets, as information is discarded and it cannot be recovered. Between the two approaches we have cancelable biometrics that apply irreversible transformations to protect biometric data, whose assurances of irreversibility rely on the difficulty to obtain good approximations of the original biometric data.

\subsubsection{Unlinkability}
Compared to irreversibility, the evaluation of unlinkability has received less attention in the literature, and no metrics have yet been specified by the ISO/IEC 30136. Common approaches consist in the definition of linkage functions to determine if multiple representations of protected biometric data belong to the same individual, and the consequent evaluation of these functions with traditional performance metrics, such as FMR and FNMR, %and equal error rate (EER), 
to compare the performances obtained when biometric data are protected with the same or different keys \cite{Zhou2012PrivacyAS}. %buhan_quantitative_2009
This allows to assess if comparison of biometric data protected with different keys is at least as hard as achieving false matches.% \cite{piciucco_cancelable_2016}. 
Other empirical evaluations of unlinkability have been proposed in \cite{simoens_privacy_2009}, with heuristic that exploit the information leaked by protected biometric data, and in \cite{nagar_biometric_2010}, with attackers that match reversed biometric data.

A general framework for the evaluation of unlinkability has been proposed in \cite{gomez-barrero_general_2018}, with the definition of two metrics for the quantitative measurement of unlinkability. The first metric is \emph{score-wise} and represents the difference between the conditional probabilities of having cross-matching and non-cross-matching data given a specific similarity score $s$. It indicates if B-PETs fail to provide unlinkability for specific values of $s$. The second metric is \emph{global} and assesses in the entire score domain if the score distributions of cross-matching and non-cross-matching data overlap. The evaluation of unlinkability depends on the linkage function considered, as inaccurate functions fail to reveal threats of specific attacks. For instance, if biometric data are protected with permutations, linkage functions not only have to consider the inversion of permutations, but also attacks computing simple statistics of protected data to link individuals.

Finally, we observe that bad sources of randomness may prevent B-PETs from providing unlinkability. In cancelable biometrics the protected biometric data of the same individual require distant transformation parameters to be unlinkable, limiting the parameter space suitable for transformations \cite{rathgeb_survey_2011}. Biometric cryptosystems may require opportune randomness to hide the information about individuals that may be contained in helper data. B-PETs fail to provide unlinkability if they do not use random keys or parameters. As in the case of irreversibility, unlinkability relies on the secrecy of keys and the difficulty to obtain good approximations of the original biometric data.

\subsubsection{Privacy of soft-biometrics}

B-PETs that prevent the extraction of soft-biometric attributes from biometric data are typically provided without any formal evaluation of the proposed techniques. To measure the validity of B-PETs, classifiers of soft-biometric attributes %trained with the original biometric data
are applied to both original and protected biometric data, and performance differences are reported \cite{meden_privacyenhancing_2021}. However, these approaches assume that attackers have limited resources, but attackers that possess a database of protected biometric data labelled according to soft-biometric attributes can also train classifiers in the protected domain. Additionally, attackers may derive soft-biometric attributes when they attempt to revert or link protected biometric data, and when they observe the similarity scores obtained for non-mated samples. In particular, facial recognition systems produce higher similarity scores and consequently more false matches for individuals with similar soft-biometric attributes. An attack that successfully exploit this effect to derive soft-biometric attributes of protected biometric data is presented in \cite{osorio-roig_attack_2022}. The study encourages to consider the proposed attack in the evaluation of B-PETs. When evaluating the suppression of soft-biometric attributes, it is also important to analyse if the recognition performance achieved with protected biometric data gets worse, as soft-biometric attributes generally facilitate the recognition of individuals.
% The unstructured nature of biometric data makes it difficult to identify which portions of data are responsible for the disclosing of soft-biometric attributes.

A standardised protocol to evaluate the privacy of soft-biometric attributes has been proposed in \cite{terhorst_privacy_2020}, considering the most critical scenario of attackers that know and adapt to B-PETs. They are able to reproduce B-PETs and train an extensive set of soft-biometric classifiers with both protected and unprotected biometric data. According to the study, this attack scenario requires more consideration than others, for instance the manual investigation of the biometric data reconstructed from protected data, because the patterns of soft-biometric attributes should be easily detectable with multiple classifiers, also trained in the protected domain. Recognition performance and estimation of soft-biometric attributes are evaluated with protected and unprotected biometric data and suitable metrics. Subsequently, they are combined in the \emph{privacy gain identity loss coefficient} (PIC), that weights the gain in privacy against the loss in recognition to determine the benefit of using the analysed B-PETs. The soundness of PIC relies on the metrics used to quantify the recognition performance and the estimation of soft-biometric attributes.

While some B-PETs have been specifically designed to protect soft-biometric attributes, and they succeed even when multiple classifiers are trained with protected biometric data \cite{morales_sensitivenets_2021}, it can be assumed that cancelable biometrics, biometric cryptosystems, and homomorphic encryption achieve soft-biometric protection only when secrets are unknown to attackers.

\subsection{Evaluation of additional requirements}
Recognition accuracy degradation is a common issue of B-PETs, as the transformations applied to biometric data intend to protect privacy and not to increase their ability to distinguish individuals. Ideally B-PETs should retain the recognition performance of the original recognition systems, but it is challenging to design data transformations that achieve it and at the same time satisfy privacy requirements \cite{nandakumar_biometric_2015}. In cancelable biometrics non-invertible transformations reduce the discriminability of biometric data, while in biometric cryptosystems the use of error correction schemes precludes the design of recognition systems with sophisticated comparators. On the other side, most of the B-PETs designed to prevent the extraction of soft-biometric attributes consist in deep neural networks trained to maintain the recognition performance \cite{morales_sensitivenets_2021, bortolato_learning_2020}. Accuracy degradation is evaluated by comparing the recognition performance achieved with original and protected biometric data. A common metrics used for the evaluation is FNMR at fixed FMR. 

We have already mentioned that homomorphic encryption allows to perform encrypted comparisons and obtain the same results of unencrypted comparisons. However, this can be achieved only with an increase of computational costs, in contrast to PbD that requires not to affect the functionality of the system. The computational costs introduced by the other B-PETs are generally lower, and due to the implementation of cryptographic algorithms and the training of neural networks, with the latter that only affect the implementation phase of B-PETs. The increase of computational cost requires particular attention when biometric recognition systems perform identification, as the number of comparisons is equal to the number of individuals enrolled in the system. Computational complexity is evaluated with the number of operations or the run-time required to execute the algorithms of B-PETs. 

Finally, storage requirements must be considered when biometric data need to be stored in portable devices with low storage capacities or in barcodes. Storage requirements are evaluated according to the number of bits required to store the protected biometric data of an individual enrolled in the system \cite{ISO-IEC-30136}.

\section{Conclusion}\label{sec:conclusion}
The different concepts of B-PETs have been proposed throughout the past years along with varying nomenclature. This clouds the picture of the data protection landscape in the area of biometric recognition which is essential for stakeholders and practitioners in the field. 

This work summarised the properties of \emph{all} existing B-PETs within a generic framework. Different concepts of B-PETs are compared in detail at each processing step and their main properties are described and related to data protection techniques and principles. The latter is of particular interest for non-experts and facilitates an effective choice of B-PETs depending on application scenarios and data protection requirements. Eventually, basic approaches for the evaluation of key properties and additional practical requirements of B-PETs are summarised.

% or


%\appendices

%\section{}

\ifCLASSOPTIONcompsoc
  \section*{Acknowledgments}
\else
  \section*{Acknowledgment}
\fi

This work has in part received funding from the European Union’s Horizon 2020  research and innovation programme under the Marie Skłodowska-Curie grant agreement No. 860813 - TReSPAsS-ETN and the German Federal Ministry of Education and Research and the Hessen State Ministry for Higher Education, Research and the Arts within their joint support of the National Research Center for Applied Cybersecurity ATHENE. R. Tolosana and R. Vera-Rodriguez are also supported by INTER-ACTION (PID2021-126521OB-I00 MICINN/FEDER).

\ifCLASSOPTIONcaptionsoff
  \newpage
\fi



\bibliographystyle{IEEEtran}
\bibliography{reference}
 
\begin{IEEEbiography}[{\includegraphics[width=1in,height=1.25in,clip,keepaspectratio]{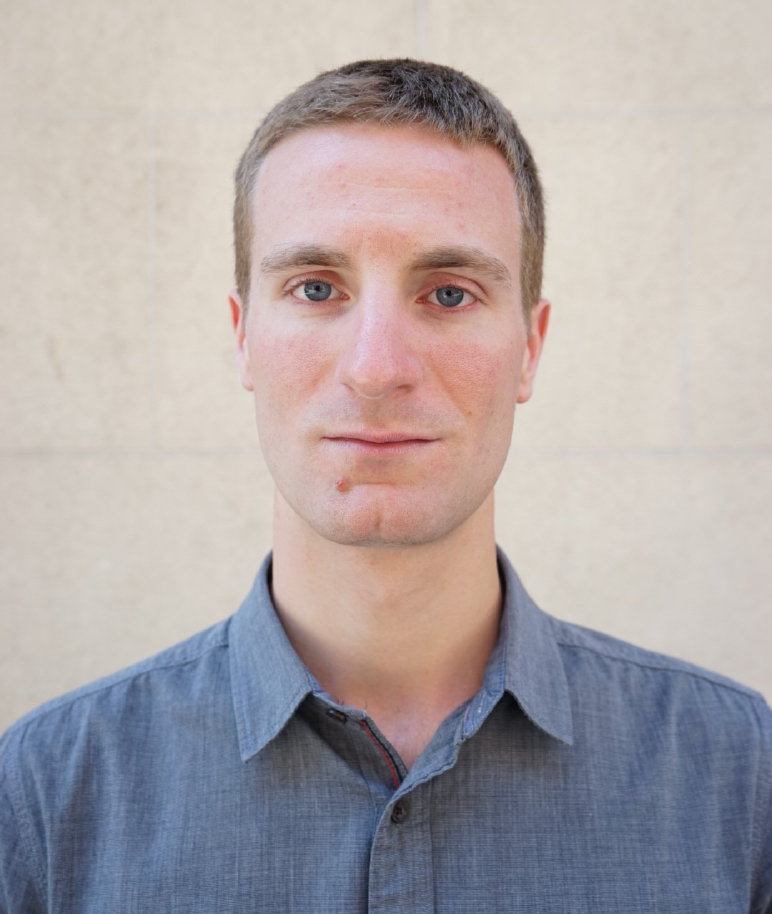}}]{Pietro Melzi}
received his B.Sc. degree in Engineering of Computer Systems from Politecnico di Milano, Italy, in 2017, and his M.Sc. degree in Computer Science and Engineering from Politecnico di Milano, Italy, in 2020. Since 2020, he is PhD student at Universidad Autonoma de Madrid. His research interests include human-computer interaction, artificial intelligence for healthcare, biometrics, and privacy aspects of biometric systems.

\end{IEEEbiography}

\begin{IEEEbiography}[{\includegraphics[width=1in,height=1.25in,clip,keepaspectratio]{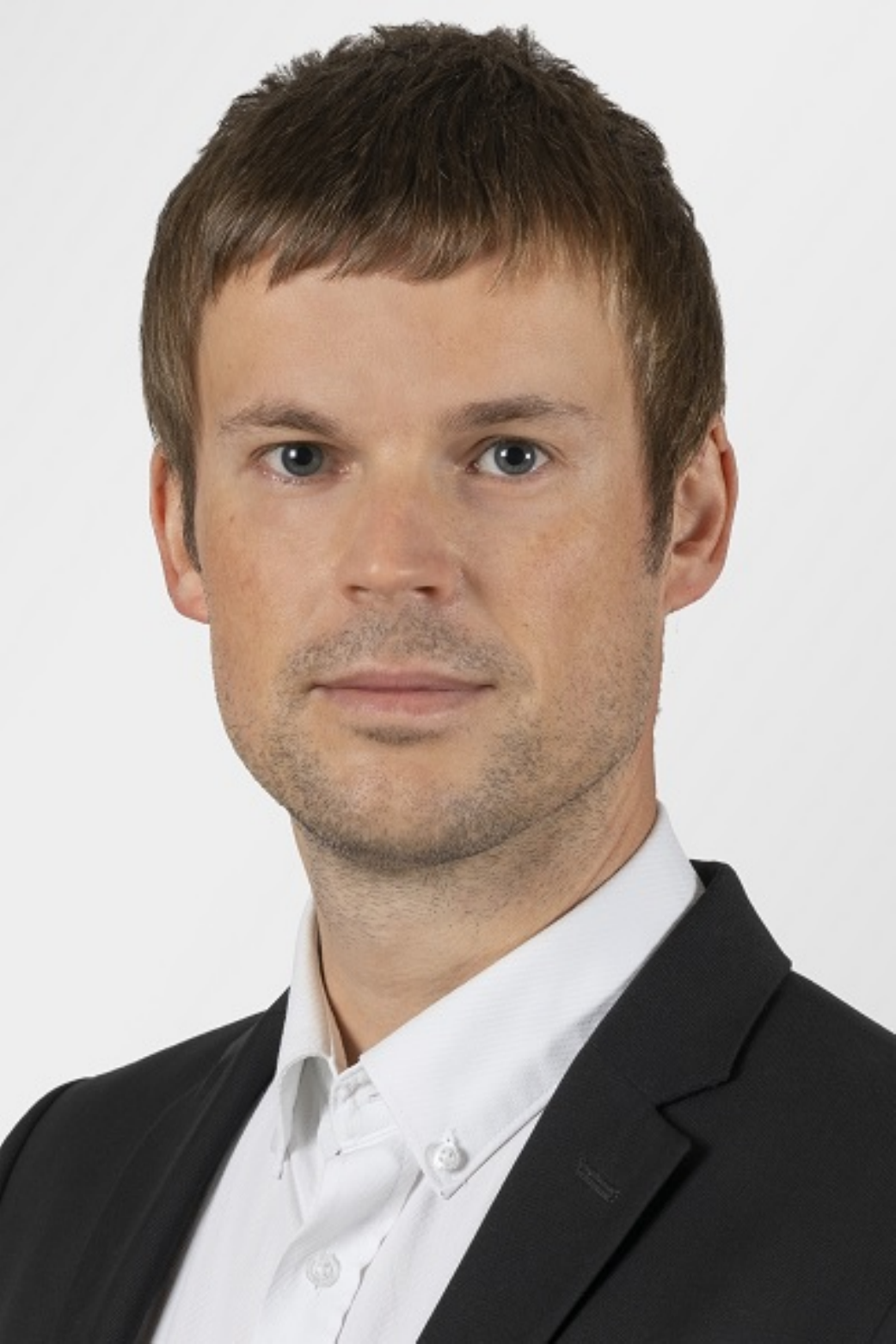}}]{Dr. Christian Rathgeb}
is a Senior Researcher with the Faculty of Computer Science, Hochschule Darmstadt (HDA), Germany. He is a Principal Investigator in the National Research Center for Applied Cybersecurity ATHENE. His research includes pattern recognition, iris and face recognition, security aspects of biometric systems, secure process design and privacy enhancing technologies for biometric systems. He co-authored over 100 technical papers in the field of biometrics. He is a winner of the EAB - European Biometrics Research Award 2012, the Austrian Award of Excellence 2012, Best Poster Paper Awards (IJCB'11, IJCB'14, ICB'15), the Best Paper Award Bronze (ICB'18) and Best Paper Award (WIFS'21). He is a member of the European Association for Biometrics (EAB), a Program Chair of the International Conference of the Biometrics Special Interest Group (BIOSIG) and an editorial board member of IET Biometrics (IET BMT). He has served for various program committees and conferences (\eg ICB, IJCB, BIOSIG, IWBF) and journals  as a reviewer (\eg IEEE TIFS, IEEE TBIOM, IET BMT).
\end{IEEEbiography}

\begin{IEEEbiography}[{\includegraphics[width=1in,height=1.25in,clip,keepaspectratio]{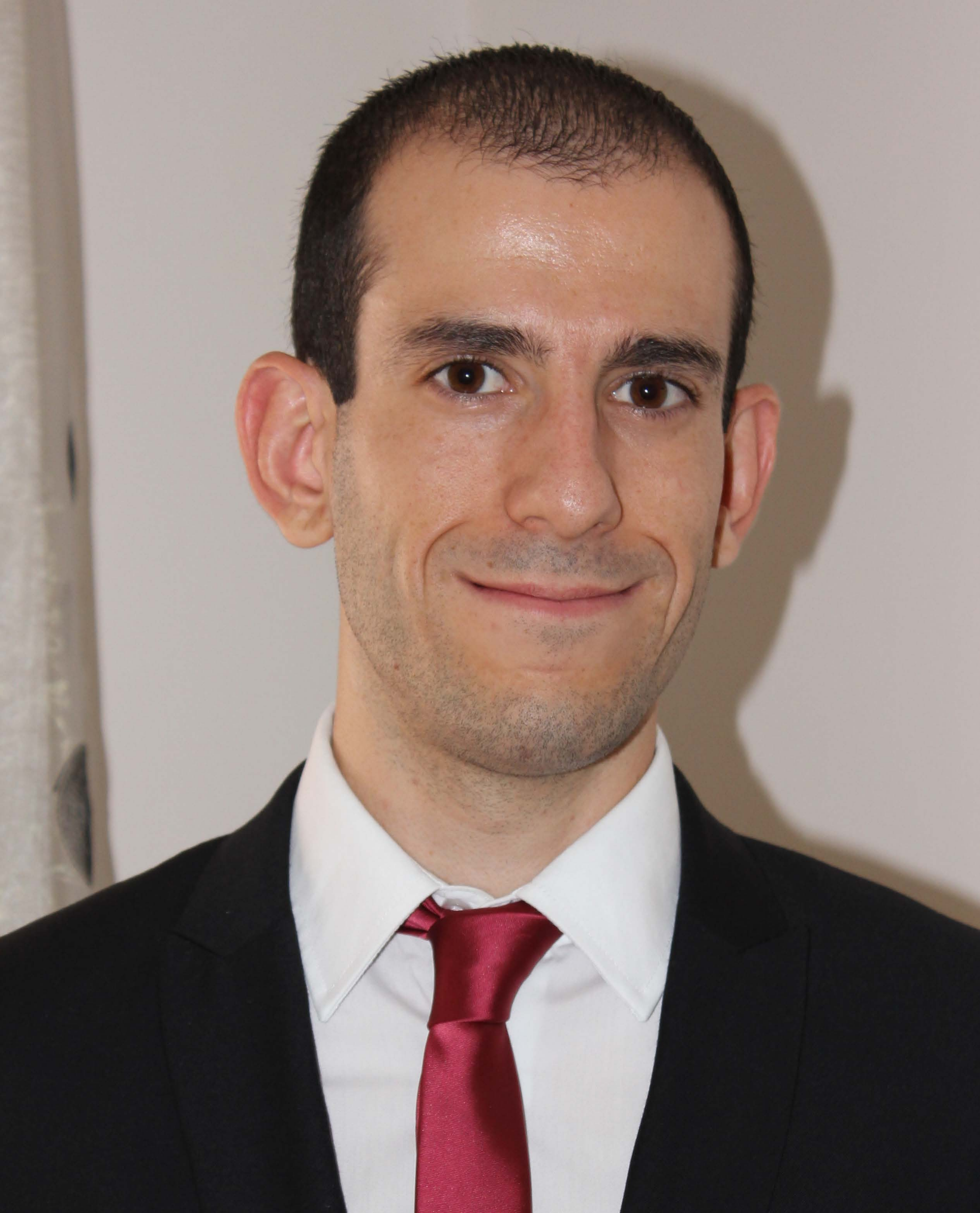}}]{Dr. Ruben Tolosana}
received the M.Sc. degree in Telecommunication Engineering, and his Ph.D. degree in Computer and Telecommunication Engineering, from Universidad Autonoma de Madrid, in 2014 and 2019, respectively. In 2014, he joined the Biometrics and Data Pattern Analytics - BiDA Lab at the Universidad Autonoma de Madrid, where he is currently collaborating as an Assistant Professor. Since then, Ruben has been granted with several awards such as the FPU research fellowship from Spanish MECD (2015), and the European Biometrics Industry Award (2018). His research interests are mainly focused on signal and image processing, pattern recognition, and machine learning, particularly in the areas of DeepFakes, HCI, and Biometrics. He is author of several publications and also collaborates as a reviewer in high-impact conferences (WACV, ICPR, ICDAR, IJCB, etc.) and journals (IEEE TPAMI, TCYB, TIFS, TIP, ACM CSUR, etc.). Finally, he is also actively involved in several National and European projects.
\end{IEEEbiography}

\begin{IEEEbiography}[{\includegraphics[width=1in,height=1.25in,clip,keepaspectratio]{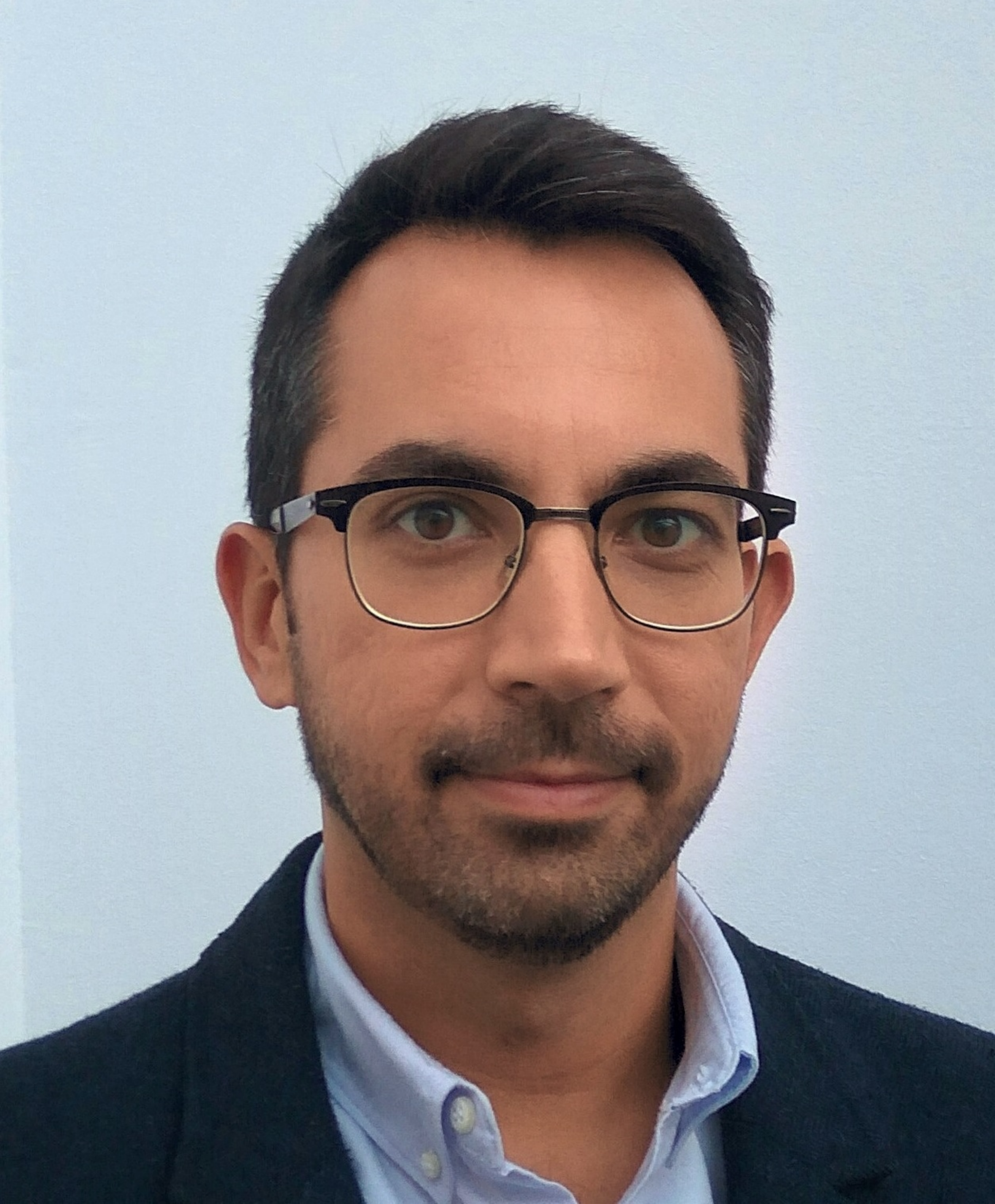}}]{Prof. Dr. Ruben Vera/Rodriguez}
received the M.Sc. degree in telecommunications engineering from Universidad de Sevilla, Spain, in 2006, and the Ph.D. degree in electrical and electronic engineering from Swansea University, U.K., in 2010. Since 2010, he has been affiliated with the Biometric Recognition Group, Universidad Autonoma de Madrid, Spain,
where he is currently an Associate Professor since 2018. His research interests include signal and image
processing, pattern recognition, HCI, and biometrics, with emphasis on signature, face, gait verification
and forensic applications of biometrics. Ruben has published over 100 scientific articles published in international journals and conferences. He is actively involved in several National and European projects focused on biometrics. Ruben has been Program Chair for the IEEE 51st International Carnahan Conference on Security and Technology (ICCST) in 2017; the 23rd Iberoamerican Congress on Pattern Recognition (CIARP 2018) in 2018; and the International Conference on Biometric Engineering and Applications (ICBEA 2019) in 2019.
\end{IEEEbiography}

\begin{IEEEbiography}[{\includegraphics[width=1in,height=1.25in,clip,keepaspectratio]{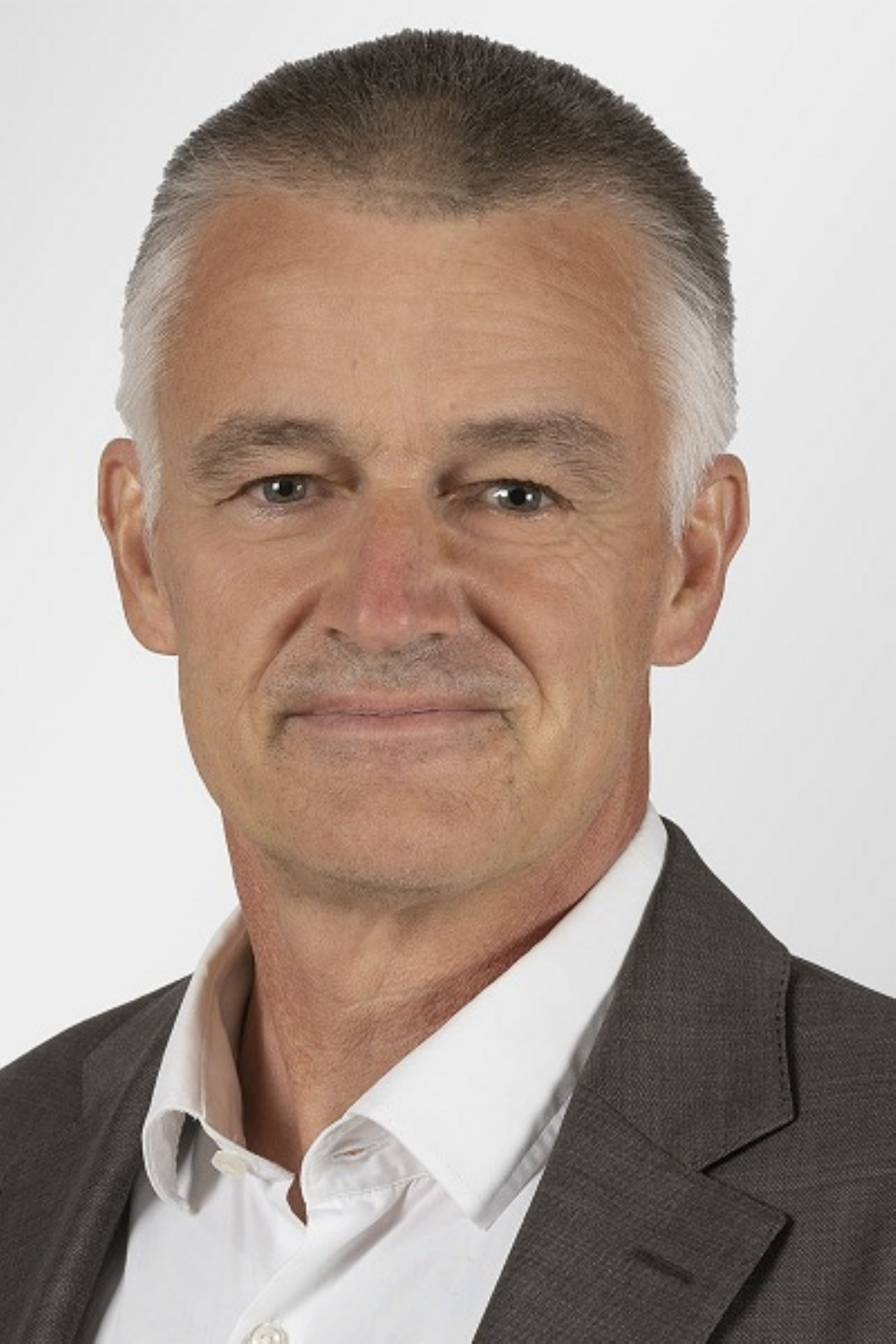}}]{Prof. Dr. Christoph Busch} is member of the Norwegian University of Science and Technology (NTNU), Norway. He holds a joint appointment with Hochschule Darmstadt (HDA), Germany. Further he lectures Biometric Systems at Denmark’s DTU since 2007. On behalf of the German BSI he has been the coordinator for the project series BioIS, BioFace, BioFinger, BioKeyS Pilot-DB, KBEinweg and NFIQ2.0. He was/is partner of the EU projects 3D-Face, FIDELITY, TURBINE, SOTAMD, RESPECT, TReSPsS, iMARS and others. He is also principal investigator in the German National Research Center for Applied Cybersecurity (ATHENE) and is co-founder of the European Association for Biometrics (EAB). Christoph co-authored more than 500 technical papers and has been a speaker at international conferences. He is member of the editorial board of the IET journal on Biometrics and formerly of the IEEE TIFS journal. Furthermore, he chairs the TeleTrusT biometrics working group as well as the German standardization body on Biometrics and is convenor of WG3 in ISO/IEC JTC1 SC37.
\end{IEEEbiography}

\end{document}